# Optimising Call Centre Operations using Reinforcement Learning: Value Iteration versus Proximal Policy Optimisation

Kwong Ho LI [1,*] and Wathsala Karunarathne [1]

[1] *School of Computer and Mathematical Sciences, Faculty of Sciences, Engineering and Technology (SET), The University of Adelaide, North Terrace, Adelaide, 5005, Australia*

[*]*Corresponding author: kwongho.li@gmail.com*

**Abstract.** This paper investigates the application of Reinforcement Learning (RL) to optimise call routing in call centres to minimise client waiting time and staff idle time. Two methods are compared: a model-based approach using Value Iteration (VI) under known system dynamics, and a model-free approach using Proximal Policy Optimisation (PPO) that learns from experience. For the model-based approach, a theoretical model is used, while a simulation model combining Discrete Event Simulation (DES) with the OpenAI Gym environment is developed for model-free learning. Both models frame the problem as a Markov Decision Process (MDP) within a Skills-Based Routing (SBR) framework, with Poisson client arrivals and exponentially distributed service and abandonment times. For policy evaluation, random, VI, and PPO policies are evaluated using the simulation model. After 1,000 test episodes, PPO consistently achives the highest rewards, along with the lowest client waiting time and staff idle time, despite requiring longer training time.

Keywords: Reinforcement learning, Discrete event simulation, Skills-based routing, Markov Decision Process, Policy optimisations.

## INTRODUCTION

Optimal task assignment and resource sharing are critical in modern service systems like call centres [1]. This paper explores RL techniques for improving call routing policy, comparing model-based and model-free methods in terms of practicality and performance. The objective is to minimise client waiting time and staff idle time through adaptive decisions.

The motivation for this study is the limitation of model-based methods, which require full knowledge of transition probabilities and reward functions which is an assumption rarely met in real-world settings. Model-based methods face significant challenges in real-world applications due to the complexity and uncertainty of actual environments, where accurately modeling environment dynamics is often impractical [2]. In contrast, model-free methods learn directly from interaction, using sampled trajectories to estimate value functions or optimise policies without relying on known dynamics. Agents using model-free approaches can learn effective policies by updating through direct interaction, without requiring explicit transition models [3].

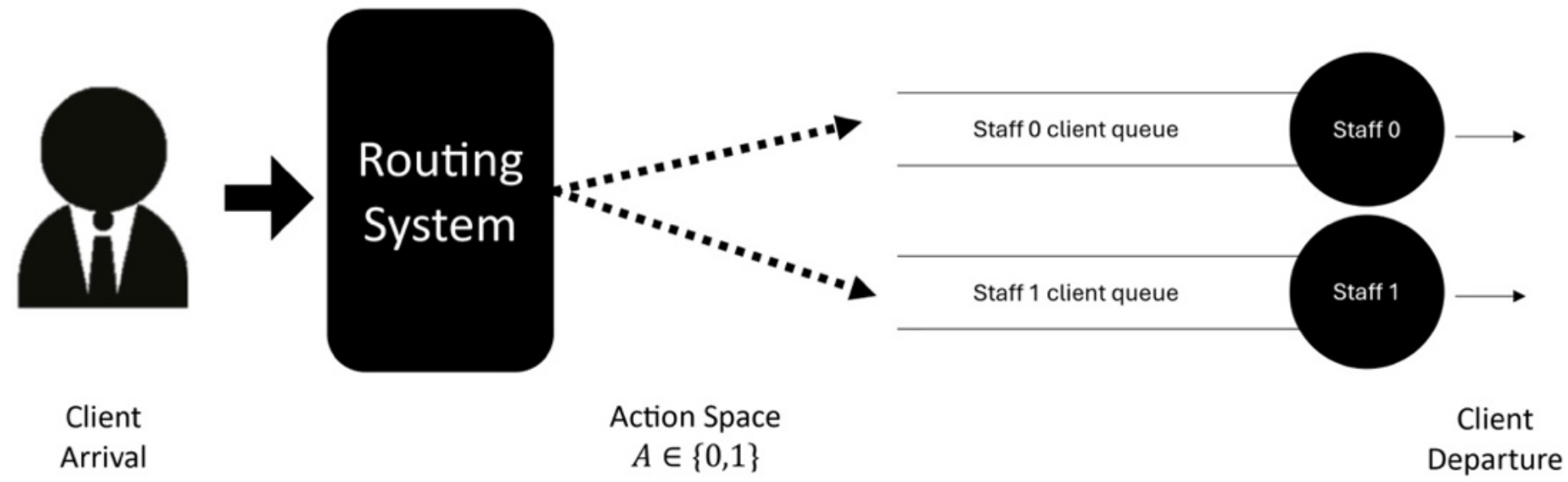

**Figure 1.** Figure showing call centre environment setting used in this study. Clients arrive stochastically and are routed by a central decision-making system to one of two staff-specific queues based on the action. Staff 0 and Staff 1 serve clients according to their respective queues. After service completion, clients exit the system.

The call centre simulates an 8-hour (28,800-second) workday, a typical customer service setup. Clients are routed immediately to staff upon calling, and each staff member has their own First-In-First-Out (FIFO) queue. Clients stay in queue until they are served or abandoned due to excessive waiting. Staff serve the next client immediately after finishing a call. Routing follows a SBR strategy, considering both queue lengths and inquiry types, as staff handle calls with varying efficiency [4]. This study considers a simplified call centre environment with two staff members (Staff 0 and Staff 1) and two kinds of inquiry types (Type 0 and Type 1). Staff can handle all inquiry types, although efficiency varies by inquiry type and staff member.

Two models are developed in this study: a theoretical model and a simulation model. Both formulate the routing problem as a MDP [5,6] with finite state and action spaces and a defined reward function. The state space includes queue lengths for two staff members and the type of incoming inquiry, while the action space consists of assigning the call to one of the staff. The model follows a SBR approach, where routing decisions are made dynamically based on the current state [4]. The reward function is designed to shape the learning behavior of routing system [5]. The call centre is modeled as events occurring randomly over time, following a Poisson process. Events occur independently at a constant average rate, are memoryless, and occur one at a time where no simultaneous events are allowed [7]. Client arrivals follow a Poisson distribution, and both service and abandonment times are both assumed to be exponentially distributed, consistent with the SBR framework [4]. The theoretical model assumes full knowledge of the environment's transition dynamics and applies a dynamic programming method, VI [8] to compute the optimal policy. In contrast, the simulation model is designed for model-free learning and policy evaluation. It integrates a DES system within the OpenAI Gym framework [9,10], enabling realistic emulation of stochastic call centre operations. Importantly, client arrivals, service times, and abandonment are drawn from probability distributions rather than using a fixed random seed, thus generating variability and realism across episodes.

Recent research has also explored the use of reinforcement learning methods to optimise queueing systems in broader contexts. For instance, Liu et al. [11] has applied VI to a class of controlled queueing networks, demonstrating that dynamic programming techniques can yield optimal policies under known environment dynamics. In parallel, Dai and Gluzman [12] has implemented deep reinforcement learning using PPO [13] to control complex queueing networks, showing that model-free approaches can effectively manage large-scale, stochastic service systems. These studies reinforce the applicability of both model-based and model-free methods to queuing system optimisation problems.

To evaluate performance, three policies are tested within the simulation model: a random policy as a baseline, the VI policy from the theoretical model, and a model-free policy learned using PPO. All policies are evaluated over 1,000 simulation episodes. PPO achieves the highest overall reward and outperforms both baselines in reducing client waiting time and staff idle time, validating its effectiveness in complex, uncertain environments. However, it requires more training time compared to VI.

# MATERIALS AND METHODS

## Materials

The simulation environment was developed using Python version 3.8 within Visual Studio Code as the integrated development environment. All experiments were conducted on an Apple MacBook Air powered by an Apple M1 chip, equipped with 8 GB of RAM, running macOS version 15.

The total simulation time per episode was set to 28,800 seconds. Key simulation parameters are summarised in Table 1. These include inter-arrival times, abandonment, and service times for two inquiry types and two staff roles. The configuration reflects a skills-based routing environment, where Type 0 inquiries are more frequent and more prone to abandonment, while Staff 0 specialises in handling Type 0 inquiries, Staff 1 is better suited for Type 1.To find a reliable evaluation length, exploratory runs with random routing and no fixed seed were conducted to tune key parameters. Based on 1,000 simulation runs, the average number of processed clients stabilised, consistent with the Law of Large Numbers, ensuring reliable evaluation metrics [14]. During these runs, queue lengths occasionally reached a maximum of 10 for both staff members. To prevent truncation effects and ensure adequate buffer capacity, the maximum queue length was conservatively set to 14.

**Table 1.** Mean parameter values of call centre configuration. Type 0 inquiries arrive more frequently and have a higher likelihood of abandonment compared to Type 1. Staff 0 is more efficient at handling Type 0 inquiries, while Staff 1 is more suited to Type 1 inquiries, reflecting a skill-based routing structure.

| | Type 0 | Type 1 |
|---|---|---|
| Inter-arrival time | 100 | 120 |
| Abandonment | 300 | 400 |
| Staff 0 Service | 120 | 190 |
| Staff 1 Service | 150 | 170 |

## The Call Centre Environment

In this research, both theoretical model and simulation model frame the call routing problem as a MDP. This shared formulation allows consistent modeling of the call centre's configuration, operational assumptions, and the definition of state and action spaces across both approaches. This call centre environment's design follows the SBR strategy, which considers both current queue lengths and the type of inquiry [4]. The primary difference between the two models lies in the implementation of the environment and the design of the reward function, the theoretical model uses a deterministic, fully-known transition system, while the simulation model incorporates stochastic, event-driven dynamics to reflect real-world uncertainties.

An MDP is a mathematical framework used to model decision-making in environments with both randomness and control. It consists of five components: a finite set of states $S$, a set of actions $A$, a transition probability function $P(s' \mid s, a)$, a reward function $\mathrm{R(s,a)}$, and a discount factor $\gamma \in [0,1]$ that weights future rewards [6]. In our call centre routing system, the agent (the routing mechanism) selects actions based only on the current observable state. This satisfies the Markov property, meaning the system's future state depends only on its present state and action, not on past states. This makes the MDP a suitable and standard approach in reinforcement learning for such dynamic environments.

The environment's state is defined by three variables: the number of clients in Staff 0's queue ($n_0$), the number in Staff 1's queue ($n_1$), and the type of inquiry being routed ($\tau \in 0{,}1$). Formally, the state space is $S = (n_0, n_1, \tau)$. The action space is discrete, with the system choosing to route the incoming client to either Staff 0 or Staff 1, represented as $a \in 0{,}1$, where $a = 0$ assigns the client to Staff 0 and $a = 1$ assigns the client to Staff 1.

The assumptions underpinning this environment are as follows:

- All inquiries are assumed to be classified upon arrival.
- All clients must pass through the routing system and the client queue before receiving service.
- Clients are routed at arrival and remain in their assigned queue until served or they abandon.
- Queue switching is not allowed.
- Staff immediately serve the next client after completing a call.
- Clients entering before the end of the working hours are eventually serviced, even if staff work overtime.
- The system operates under stationary condition, with no peak time or time-of-day effects.
- Staff members do not take breaks and are immediately available for the next client once a call is completed.

## The Theoretical Model

VI is a classic model-based RL method for computing the optimal policy in a MDP, under the assumption that all transition probabilities are fully known. The theoretical model is built to apply VI to the call routing problem. Since the probability of two simultaneous Poisson events is infinitesimal, it is reasonable to assume that only one event occurs at any given moment in the system. This simplifies the environment into a deterministic transition model, for a given state-action pair $(\boldsymbol{s}, \boldsymbol{a})$, there is only one possible next state $\boldsymbol{s'}$. The probability of the next state would be one. The reward function is designed to minimise client waiting time and discourage idle staff. The function is defined as:

$$R(s,a)\begin{cases} -158, & \text{If assigned to a full queue} \\ -158, & \text{If assigned to a busy staff while the other is idle.} \\ -1 \times E[\text{Waiting time}], & \text{If assigned staff has client waiting.} \end{cases} \quad (1)$$

**Equation 1.** Reward function used in the theoretical model. The function penalises suboptimal routing decisions to discourage assigning clients to full queues or inefficiently utilising staff. A penalty of $-158$ is applied when assigning to a full queue or a busy staff while the other is idle. Otherwise, the reward is the negative expected client waiting time, promoting actions that minimise delays.

To compute the expected waiting time, the model uses the average service rates of the two staff members and the number of clients waiting. If a staff member's queue is empty, the state is considered idle. The $-158$ penalty constant comes from the average of both staff members' mean service times: $\mathbf{(155+160)/2 = 157.5 \approx 158}$. This value is used to penalise poor routing decisions such as assigning a client to a full queue or to a busy staff member while the other is idle. Using a common penalty ensures symmetry and fairness in evaluating such decisions, regardless of staff identity. While inquiry type does not affect service time in these penalty cases as only one event is assumed to happen at a time, it is still part of the state representation. This enables more consistent comparisons between states with other policies.

## The Simulation Model

To evaluate policy performance and compare model-free RL algorithms in settings with unknown transition dynamics, a custom simulation environment was developed by integrating DES with the OpenAI Gym framework [9,10]. The hybrid method allows for both realistic modeling and experimentation with RL. The environment emulates a realistic call centre system, the situation with stochastic client arrivals, service processes, and abandonment behavior. Clients are dynamically generated during the simulation, and all underlying events are governed by probabilistic distributions. No fixed random seed is used during training, with variability and improving robustness through exposure to diverse scenarios. Unlike model-based approaches, that rely on known transition probabilities, the environment is tailored for model-free methods which is PPO. Agents learn optimal routing strategies by interacting with the environment, and therefore this is a realistic and challenging testbed for RL in service systems.

The simulation is built upon the principles of DES, a mathematical modeling approach suitable for systems where events occur at discrete points in time. The DES system consists of two main components: the event itself and the event queue [15]. The event queue, implemented as a priority queue, stores all future events that have been scheduled but not yet executed. Each event is associated with a timestamp, which determines its position in the queue. Events are inserted into the queue based on samples drawn from the relevant statistical distribution, and their timestamps are calculated by cumulatively adding the sampled time to the current simulation time. Events are processed in chronological order, and the simulation's global clock is advanced to the time of the next event. Upon execution, activities will occur and a new event may be scheduled into event queue. The DES continues processing events until the event queue is empty, effectively simulating a full day of call centre operations. There are three types of events in the simulation:

- **Arrival:** Triggered when a client contacts the call centre. Upon arrival, the client is assigned to a staff member's queue based on the action taken by the routing system. An abandonment event is simultaneously scheduled for this client, representing the possibility of the client hanging up due to long wait times. If the next arrival time has not reached the closing time, the next arrival for the same inquiry type is also scheduled, ensuring that no new calls are processed beyond working hours. If the assigned staff is available at the time of the event, the client immediately begins service, their abandonment event is canceled, and a departure event is scheduled based on a sampled service time. If the assigned staff is not available, the client remains in the queue and waits for their turn.

- **Departure:** Marks the completion of a client's service. When it occurs, the client is removed from the staff's current task. If additional clients are waiting in that staff member's queue, the next client is moved to service, their abandonment event is removed, and a new departure event is scheduled. If no clients are waiting, the staff member becomes idle, and the simulation awaits the next event.

- **Abandonment:** Occurs when a client who has been queued too long decides to leave. The client is removed from the staff's waiting queue, and no further action is taken.

The DES system is wrapped within an OpenAI Gym environment, allowing RL agents to interact with it through a standardised framework. The Gym interface encapsulates both the DES system and the staff pool which is the collection of service agents, while managing the transition logic and observation state structure. This integration enables model-free learning algorithms to perceive the environment as a MDP [16], despite the lack of an explicit transition model. The Gym environment handles the environment reset, step execution, reward calculation, and state transitions in coordination with the underlying DES logic.

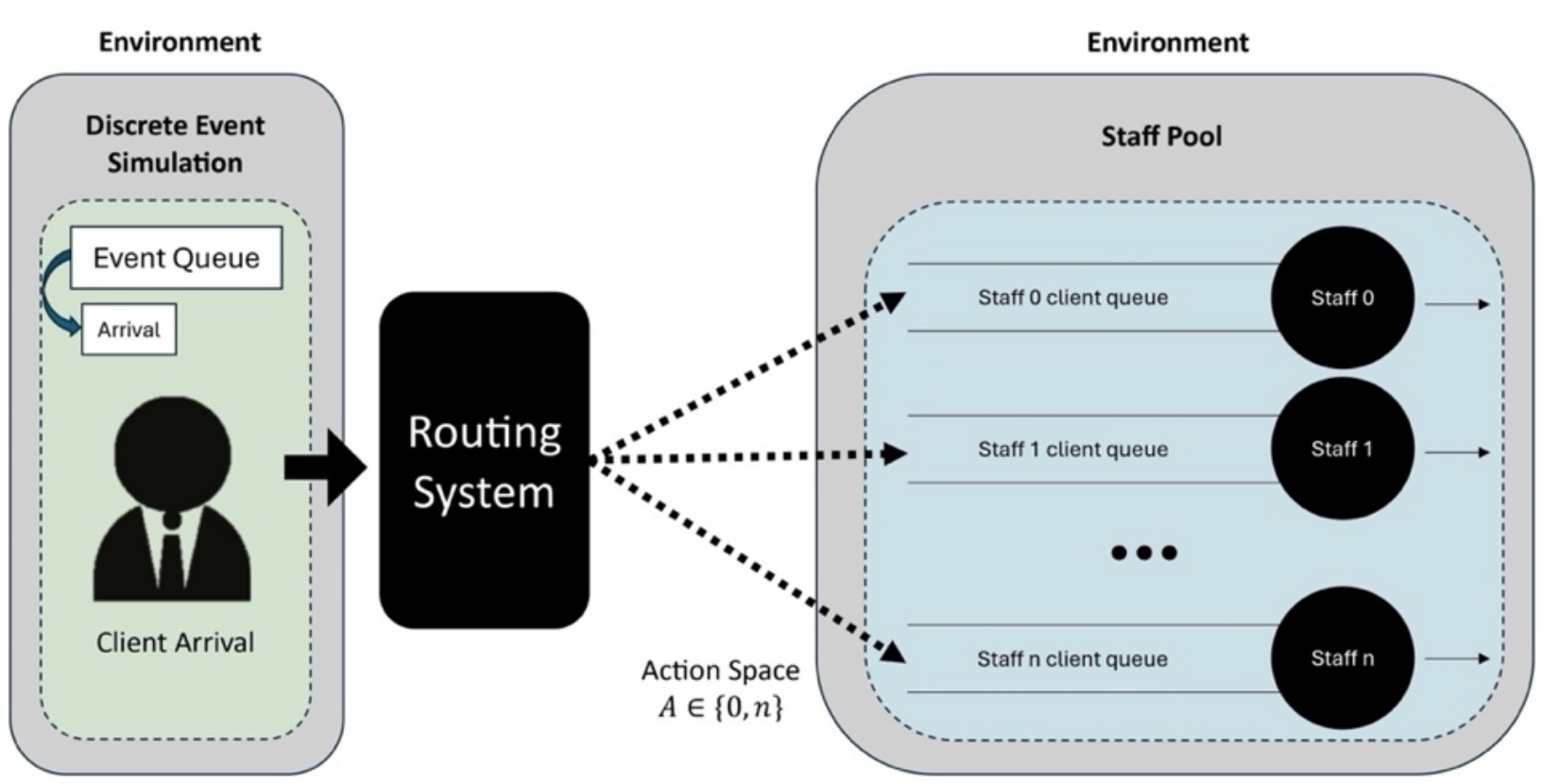

**Figure 2.** Flowchart showing client servicing in the simulation model, composed of three main components: the DES system, the Routing System, and the Staff Pool. Clients' arrival generates by DES, routing system decide which client is routed to, and staff will proceed to handle the client in staff pool.

The system can be represented by an $M/M/n$ queueing model, where are $n$ number of identical servers. For the call centre situation, each server represents a staff member, and the value of $n$ is defined by the staffing configuration parameters. Figure 2 illustrates that the flow of clients being served is structured by three components:

- **DES system:** It manages the generation of events and governs the progression of time throughout the simulation. The event queue is maintained by the DES system and that is the timeline of the simulation, where every event is placed in a queue and processed sequentially. The routing system is only activated in response to Arrival events.

- **Routing system:** The routing system is based on the current system state and makes assignment decisions whenever a new client arrives. These decisions determine which staff member the client is routed to.

- **Staff pool:** It manages the individual queues for each staff member and handles client servicing on a FIFO basis. While the simulation can accept any number of staff members, the current configuration assumes two.

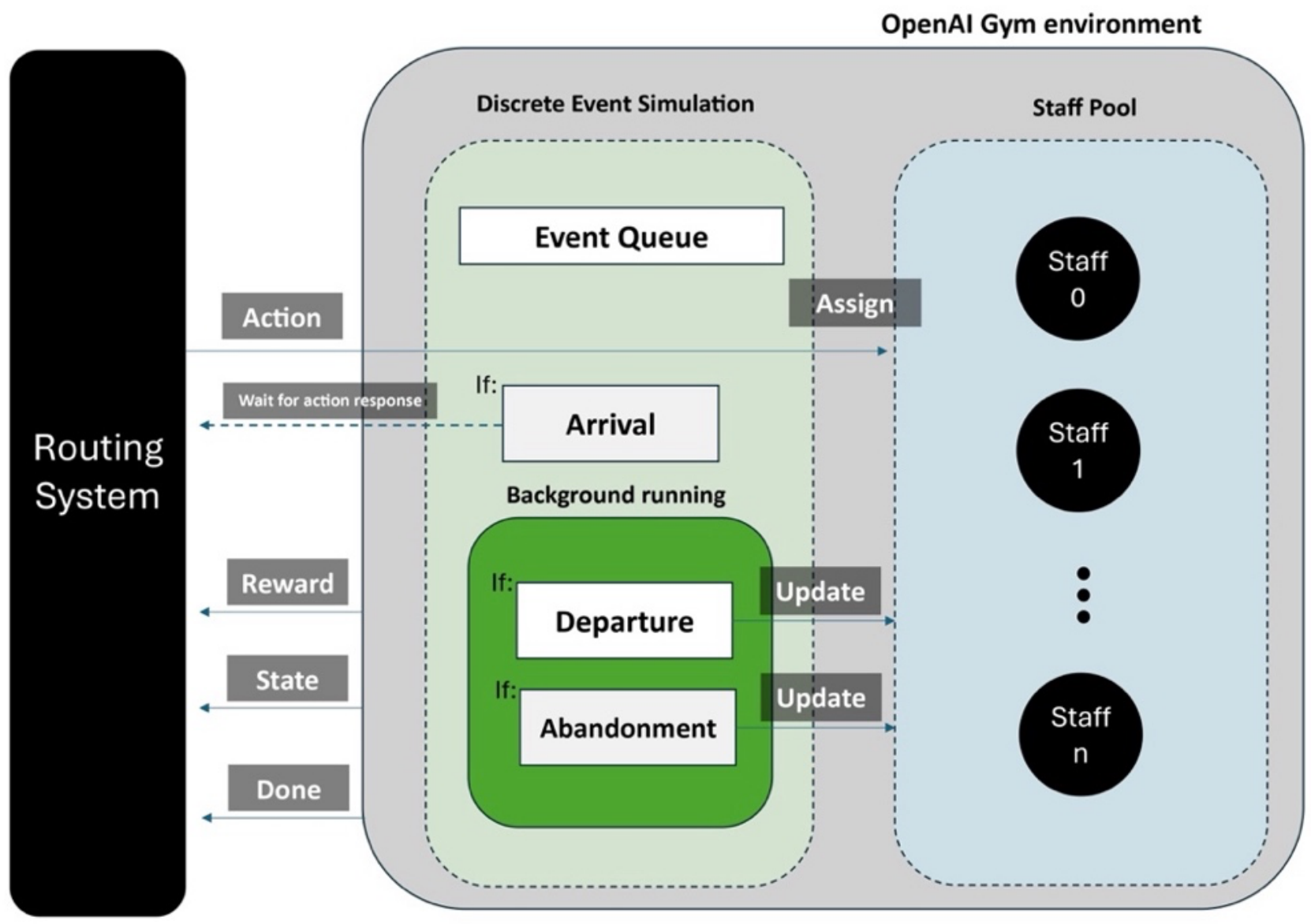

**Figure 3.** Interaction between the routing system and the environment in the simulation model. The routing system is triggered only by client arrival events, while departure and abandonment events are managed internally by the environment to update the staff pool. Upon each routing decision, the environment returns a reward, the next state, and a termination signal as feedback.

The interaction between the routing system and the simulation environment follows the classic MDP agent-environment interaction as described in the RL literature [17]. In this framework, the routing system selects an action in response to the current state, and the environment responds by returning a reward, the next state, and a termination signal as shown in Figure 3.

Upon an Arrival event, which represents a client calling the centre, the routing system is triggered to make a decision. This is the only event type that requires an action from the routing system. The chosen action assigns the incoming client to one of the staff members in the staff pool. In contrast, Departure and Abandonment events are handled automatically by the environment and update the staff pool independently in the background. These events do not require routing system intervention once a client has been routed. Any rewards associated with such background events are computed and consolidated into the reward returned during the corresponding Arrival event. This background processing still aligns with the assumption that only one event occurs at a time, events are still handled sequentially. Their timing remains independent and consistent with the underlying stochastic processes. The simulation actively responds only to events requiring routing decisions, while other events are processed passively, with the next state returned after all relevant background updates.

The reward function in the simulation environment is incentivises routing decisions that minimise both clients waiting time and staff idle time while discouraging the overloading of queues and loss of clients due to abandonment. The formal reward function is defined as:

$$R(s,a)\begin{cases} -125, & \text{Assign to full staff client queue.} \\ -1, & \text{Each second fo staff idle time during arrival event} \\ -1, & \text{Each second fo client waiting time during arrival event.} \\ -125, & \text{Each abandonment event occur.} \end{cases} \quad (2)$$

**Equation 2.** Reward function used in the simulation model. A penalty of $-125$ is applied when assigning a client to a full staff queue or when a client abandons the system. In addition, a cost of $-1$ is applied per second of staff idle time and client waiting time during each arrival event.

The structure of this reward function is grounded in observed empirical metrics and the operational goals of the call centre. The per-second penalties for staff idleness and client waiting during arrival events are designed to encourage timely and effective task allocation across the available staff. Meanwhile, the average client waiting time was approximately 126 seconds from 1,000 runs of random policy, which informed the penalty constant of $-125$ in the reward function for events such as full queues or client abandonment and serves as a deterrent against actions that cause client dissatisfaction or system inefficiency.

## RESULTS AND DISCUSSION

### Training Time

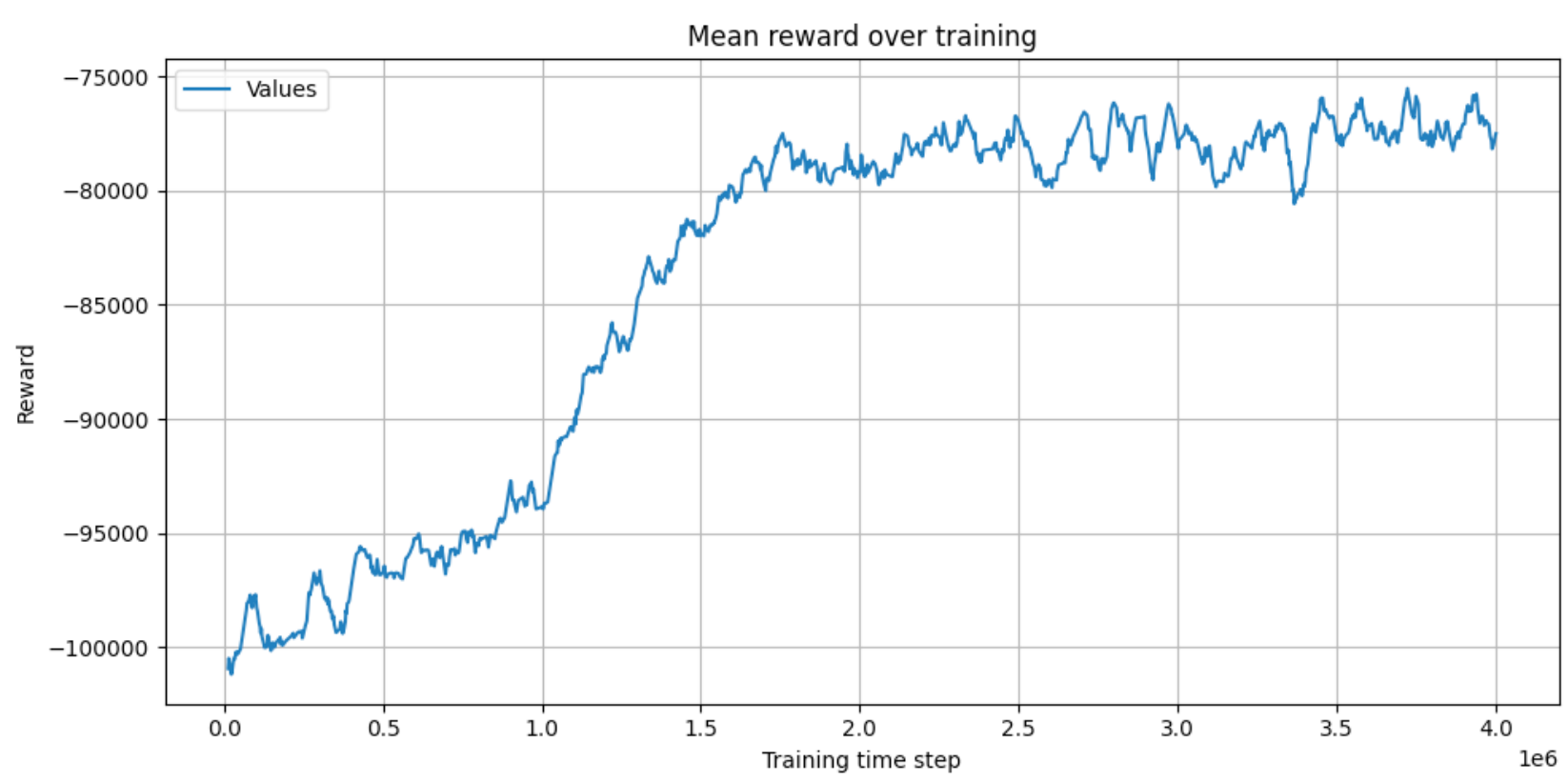

**Figure 4.** Total reward accumulated by the PPO agent over the training. The learning curve shows that the PPO algorithm begins to converge and stabilise after approximately 2 million training steps, indicating that the agent has learned an effective routing policy under the simulation environment.

Training time for both methods reflects their differences in complexity and data requirement. VI, with full access to the environment's transition dynamics, converged quickly in 205 iterations and only took 0.12 seconds. PPO, by learning trial-and-error exploration of a simulated stochastic environment, learns much more slowly. It takes about 40 minutes (2,475 seconds). As seen from Figure 4, the reward curve indicates that PPO's performance plateaued around 2 million timesteps, about halfway through the entire training process.

## Policy Evaluation

**Table 2.** Evaluation results based on 1,000 simulation runs comparing different routing policies. The best PPO policy demonstrates the best overall performance.

| | Random | Value Iteration | PPO |
|---|---|---|---|
| Total Reward | -98,900 | -84,742 | **-76,892** |
| Client Served | 322 | 344 | **366** |
| Client Abandonment | 207 | 185 | **163** |
| Client Waiting time | 127 | 114 | **98** |
| Staff 0 idle time | 4,078 | **0** | 1,064 |
| Staff 1 idle time | 3,771 | 3,600 | **968** |

Table 2 shows the results of the evaluation, averaging over 1,000 independent runs in simulation model for each policy. The comparison includes three policies from different algorithms: Random, VI, and PPO. The results demonstrate differences in performance across a range of metrics:

- The Random policy performs the worst in all metrics. It has the lowest total reward and number of served clients, the highest number of abandonment clients, and the longest mean waiting time. In addition, both staff members have high waiting and idle times, reflecting poor system utilisation.

- The optimal policy from VI improves over the random policy. It has a better total reward, fewer abandonment cases, higher service throughput, and notably minimises idle time for Staff 0. However, its static, model-based nature leads to suboptimal utilisation of Staff 1, resulting in persistent inefficiencies and comparatively high waiting times in certain scenarios.

- The optimal policy from PPO demonstrates the best performance across most of the metrics. It achieves the highest reward and serves the most clients, has the lowest abandonment clients, and reduces mean waiting time more effectively than other techniques. While it does not minimise Staff 0's idle time as Value Iteration, it offers the best balance of idle time across both staff, highlighting its adaptive capability PPO's policy achieves the highest overall reward, implying effective resource allocation and good decision-making under uncertainty.

Overall, PPO is the most practical and effective strategy for optimising call routing in this dynamic, stochastic environment. Its ability to learn and adapt from interaction data allows it to outperform model-based and heuristic approaches that lack this flexibility.

## CONCLUSION

This study explored the optimisation of a call routing strategy using RL, with a particular focus on comparing a model-based method like VI, and a model-free method like PPO. The experimental findings support the hypothesis: model-free RL offers not only greater adaptability but also significantly more practical applicability in environments where the system dynamics are partially observed, highly variable, or fundamentally unknown.

While VI delivers theoretically optimal policies under the assumption of full knowledge of transition probabilities, this assumption rarely holds in real-world systems. In complex operational settings like call centres, the probability of transitioning from one state to another depends on a wide array of factors including unpredictable client dynamic, service rates, and abandonment variability, that are nearly impossible to model exhaustively. Accurately estimating the transition probabilities for every possible state-action pair would require impractical levels of data collection, computation, and model calibration. As a result, model-based approaches like VI are fast to compute but limited in their applicability when such full environment knowledge is unavailable.

In contrast, model-free methods like PPO learn policies directly from experience, without requiring an explicit model of the environment. PPO was able to discover effective routing strategies through repeated interaction with the simulation, adjusting to stochastic variations in arrival patterns, service rates, and abandonment. This trial-and-error process is more computationally intensive and time-consuming but leads to robust and stable policies that adapt to a wide range of system conditions. PPO also demonstrated its strength in balancing staff workloads and having better performance.

One limitation of the current setup lies in the reward function design. Since the simulation is event-driven rather than time-stepped, rewards are calculated only at arrival events. This introduces approximation errors in measuring cumulative client waiting time and staff idle time, as idle periods and waiting intervals between arrivals are not continuously updated. Moreover, the reward functions differ between the theoretical model and the simulation model due to limitations in model design, which can hinder direct result comparisons between the two approaches.

Another limitation is that the current study considers only two staff agents. While this simplification aids interpretability and computational efficiency, it does not fully reflect the complexity of real-world call centres, which typically involve many concurrent service agents. However, the simulation framework and reinforcement learning architecture, particularly PPO, are inherently scalable. To support a larger number of agents, the action space can be extended to select among $n$ staff members, and the state space can be augmented to include queue lengths and availability indicators for each agent. PPO's ability to handle high-dimensional, continuous state spaces makes it well-suited for learning effective policies in such expanded environments.

Future work could address these limitations by designing a continuous-time reward model or synchronising reward definitions across both model-based and model-free implementations. Expanding the environment to include more staff, more inquiry types, or more complex routing constraints would also allow for testing the scalability and robustness of learned policies in more realistic operational settings.

These results underscore the practical value of model-free RL as a decision-making tool in complex, dynamic systems. By reducing the dependency on complete environment modeling and enabling scalable, data-driven policy learning, model-free approaches offer a promising pathway for real-world deployment in service operations, logistics, and beyond.

All the code used in this project is available at: https://github.com/nonstopronald/Reinforcement-Learning.git